\documentclass[10pt,twocolumn,letterpaper]{article}

\usepackage[pagenumbers]{wacv} % To force page numbers, e.g. for an arXiv version

\usepackage{graphicx}
\usepackage{amsmath}
\usepackage{amssymb}
\usepackage{booktabs}
\usepackage{float}
\usepackage{epsfig}
\usepackage{bm}
\usepackage{cite}
\usepackage{booktabs}
\usepackage{caption}

\usepackage[accsupp]{axessibility}

\usepackage{color} 
\newcommand{\revised}[1]{#1}

\usepackage[pagebackref,breaklinks,colorlinks]{hyperref}

\usepackage[capitalize]{cleveref}
\crefname{section}{Sec.}{Secs.}
\Crefname{section}{Section}{Sections}
\Crefname{table}{Table}{Tables}
\crefname{table}{Tab.}{Tabs.}

\usepackage{xcolor}
\usepackage{multirow}

\def\confName{WACV}
\def\confYear{2025}

\title{BeautyBank: Encoding Facial Makeup in Latent Space}

\author{Qianwen Lu$^{1,2}$, Xingchao Yang$^{1}$, Takafumi Taketomi$^{1}$\\
$^1$CyberAgent \quad$^2$The University of Tokyo\\
{\tt\small \{lu\_qianwen\_xa, you\_koutyo, taketomi\_takafumi\}@cyberagent.co.jp}
}

\begin{document}

\twocolumn[{%
\renewcommand\twocolumn[1][]{#1}%
\maketitle
\begin{center}
    \centering
    \captionsetup{type=figure}
    \includegraphics[width=0.9\textwidth]{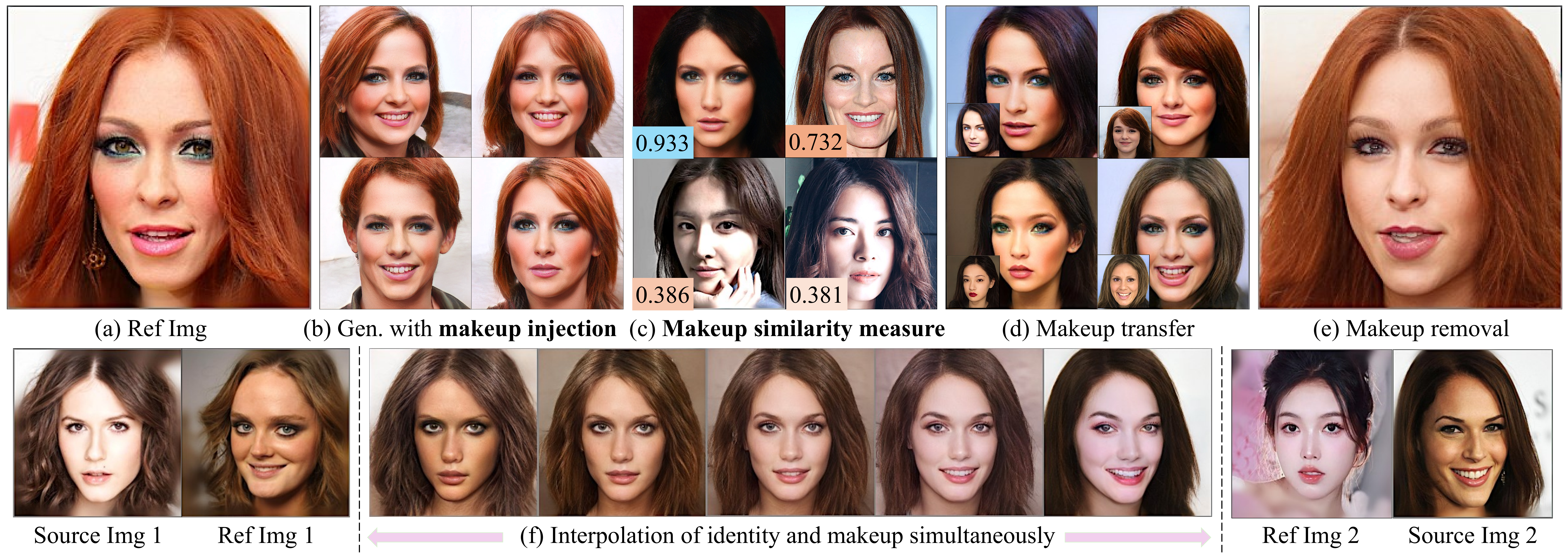}
    \captionof{figure}{\textbf{Example applications of our makeup encoder (BeautyBank).} \revised{We have successfully explored a variety of applications, including using (a) images with reference makeup to (b) generate facial images with makeup injection, (c) measure makeup similarity, and (d) transfer makeup, and (e) remove makeup. Additionally, BeautyBank can utilize two different facial identity references (Source Img 1 and 2) and two different makeup references (Ref Img 1 and 2) to (f) simultaneously interpolate identity and makeup. }The images generated using the makeup code from BeautyBank show high-quality details such as makeup colors, patterns, and textures across various makeup applications.}
    \label{fig:teaser}
\end{center}%
}]

%%%%%%%%% ABSTRACT
\begin{abstract}
    The advancement of makeup transfer, editing, and image encoding has demonstrated their effectiveness and superior quality. However, existing makeup works primarily focus on low-dimensional features such as color distributions and patterns, limiting their versatillity across a wide range of makeup applications. Futhermore, existing high-dimensional latent encoding methods mainly target global features such as structure and style, and are less effective for tasks that require detailed attention to local color and pattern features of makeup. To overcome these limitations, we propose BeautyBank, a novel makeup encoder that disentangles pattern features of bare and makeup faces. Our method encodes makeup features into a high-dimensional space, preserving essential details necessary for makeup reconstruction and broadening the scope of potential makeup research applications. We also propose a Progressive Makeup Tuning (PMT) strategy, specifically designed to enhance the preservation of detailed makeup features while preventing the inclusion of irrelevant attributes. We further explore novel makeup applications, including facial image generation with makeup injection and makeup similarity measure. Extensive empirical experiments validate that our method offers superior task adaptability and holds significant potential for widespread application in various makeup-related fields. Furthermore, to address the lack of large-scale, high-quality paired makeup datasets in the field, we constructed the Bare-Makeup Synthesis Dataset (BMS), comprising 324,000 pairs of 512x512 pixel images of bare and makeup-enhanced faces.
 \end{abstract}

 \setcounter{figure}{1}
 \begin{figure*}[t]
     \centering
     \includegraphics[width=0.76\linewidth]{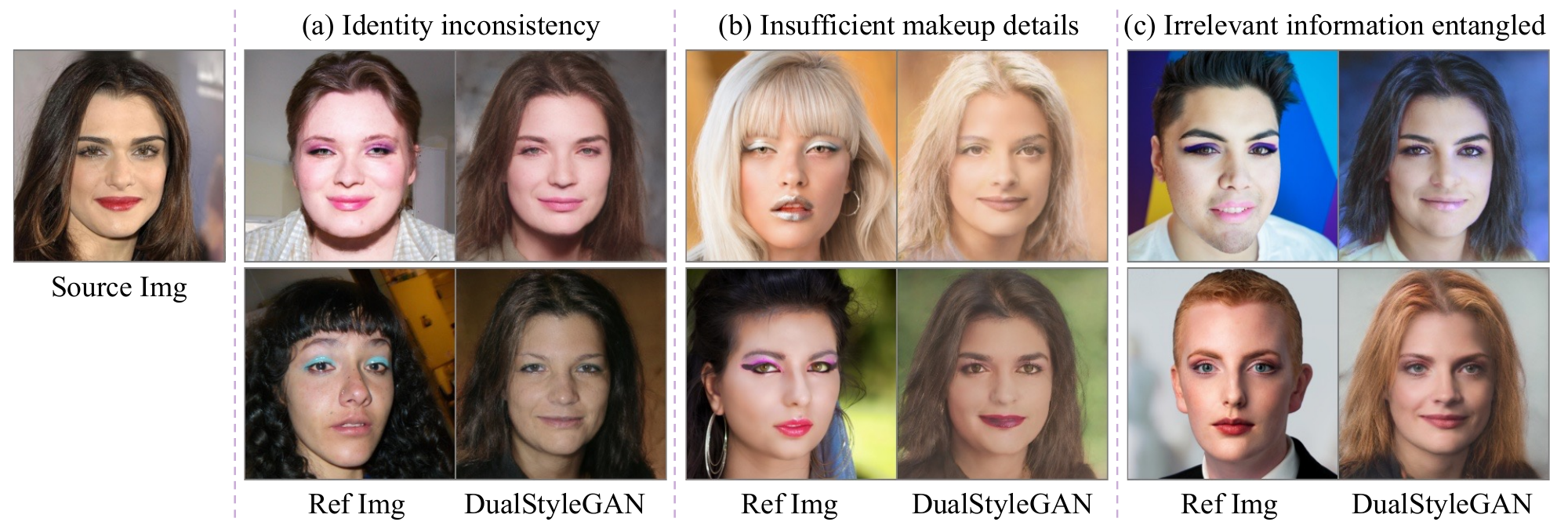}
     \caption{\textbf{Typical issues in generated images using the baseline method.} When DualStyleGAN~\cite{dualstylegan} is utilized for makeup transfer tasks, the generated images often exhibit inconsistencies in the facial identity compared to the source images. There is also a lack of detail in makeup attributes, such as local colors and patterns, and an entanglement with features that are not related to the makeup pattern.}
     \label{fig:fail_makeup}
 \end{figure*}

%%%%%%%%% BODY TEXT
\section{Introduction}
\label{sec:intro}

The rapid progress of various generative models, such as GANs and diffusion models, has significantly advanced makeup-related visual tasks\revised{~\cite{FFHQ, StyleGAN2, ho2020denoising, FaceBeautification}}.
Despite their impressive performance, the algorithms are specifically designed for certain makeup tasks, such as makeup transfer and editing~\cite{beautyGAN, PSGAN, elegant, MakeupExtract, stablemakeup}. The primary reason is that they tend to model low-dimensional representations of makeup features, such as color distributions, local details, and pattern styles~\cite{beautyGAN, CPM, elegant}. Consequently, these methods struggle to handle the diverse and intricate demands of real-world makeup applications, such as facial image generation with makeup injection and makeup similarity measure.

On the other hand, the latent code representation has shown its great performance in image generation, style transfer, and image editing~\cite{Toonify, VToonify, dualstylegan}. In paticular, these methods generate  high-quality style images by encoding high-dimensional style features and subsequently manipulating latent codes in semantically meaningful ways. It should be noted that these methods primarily focus on global features, including structural elements and overall color styles. However, makeup-related tasks emphasize the consistency of identity features between makeup and bare-face images, as well as the details of local colors and patterns in makeup. Directly applying existing methods to makeup encoding tasks can lead to significant facial identity changes or loss of local makeup details, as shown in Fig.~\ref{fig:fail_makeup} (a) and (b). Additionally, without disentangling makeup-irrelevant information, the generated images also exhibit significant alterations in non-facial areas such as hair and background, as shown in Fig.~\ref{fig:fail_makeup} (c).

In this paper, we propose a novel makeup encoding method that efficiently encodes facial makeup features into a high-dimensional latent space. Our method adapts to various makeup applications while preserving detailed information essential for high-quality makeup reconstruction. We initially introduce BeautyBank, a makeup encoder featuring separate paths for bare-face and makeup styles. During the training of the bare-face style path, we applied a facial enhancement loss to maintain the consistency of identity features in the bare-face code. The refined bare-face code can subsequently improve the makeup style path's ability to encode makeup representations independently. Additionally, we introduce a Progressive Makeup Tuning (PMT) strategy that employs varied training strategies and loss functions at different stages to progressively fine-tune the makeup code. BeautyBank achieves stable makeup encoding, preserves rich makeup detail features, and effectively disentangles unrelated features, such as hair and background, from the makeup encoding process. Furthermore, given the current lack of large-scale, high-quality paired makeup datasets, we construct the Bare-Makeup Synthesis Dataset (BMS), comprising 324,000 pairs of 512x512 pixel bare-face to makeup face images. This dataset provides a diverse array of makeup data for makeup encoding tasks. We generate the makeup data in the BMS dataset using LEDITS++~\cite{LEDITS++} based on style and color prompts collected from the FFHQ~\cite{FFHQ} dataset, encompassing a wide variety of makeup styles, colors, and patterns. 

In summary, our contributions are threefold:
\begin{itemize}
\item  We introduce BeautyBank, a novel makeup encoder that effectively disentangles bare-face features from makeup style features. This facilitates the encoding of makeup in a high-dimensional feature space. Our experiments demonstrate that our method expands the range of makeup applications beyond existing methods, enabling facial image generation with makeup injection and makeup similarity measure, as shown in Fig.~\ref{fig:teaser}.

\item We design the PMT strategy that incrementally fine-tunes makeup encoding. This strategy ensures the preservation of essential makeup detail features, such as color textures, while reducing the influence of makeup-unrelated features.

\item We construct the BMS datasaet, a large-scale, high-resolution makeup dataset that ensures diversity in makeup encoding. To our knowledge, this is the first large-scale dataset of its kind, consisting of paired 512x512 pixel images of bare and made-up faces. We will make this dataset publicly available and hope it can assist future makeup-related research.
\end{itemize}

%-------------------------------------------------------------------------

\section{Related Work}

\subsection{Facial Makeup Tasks}
Facial makeup is an important aspect of human appearance. In computer vision and graphics, mainstream research focuses on makeup transfer~\cite{MakeStar, beautyGAN, PairedCycleGAN, ladn, BeautyGlow, SCGAN, SOGAN, ca-gan, PSGAN, ssat, CPM, RamGAN, elegant, BeautyREC, CSD-MT, TinyBeauty, stablemakeup}, 3D makeup~\cite{CGF11Makeup, PBCR, SimulatingMakeup, bareskinnet, MakeupExtract, AvatarStudio, yang2024makeuppriors}, and face verification~\cite{protectmakeup, CLIP2Protect}. 

The task of makeup transfer is transferring a makeup pattern in a specified reference face image to a source face image. Early research focused on the color distribution of makeup~\cite{beautyGAN}, while more recent studies attempt to transfer complex makeup patterns~\cite{stablemakeup}. In addition, several studies have analyzed factors in facial images, which allows for makeup transfer to accommodate variations such as lighting~\cite{MakeupExtract}, occlusion~\cite{SOGAN}, and head pose~\cite{PSGAN, elegant}. However, most methods are limited to low resolutions, such as $256 \times 256$. 3D makeup research primarily focuses on the 3D makeup estimation or the beautification and stylization of avatars~\cite{clipface}. Tasks related to makeup in face verification~\cite{protectmakeup, CLIP2Protect} underscore the importance of security and face protection. They achieve this by adding makeup to faces, thereby generating images that aid in privacy protection. It's also worth noting that research dedicated specifically to makeup recommendation is somewhat limited~\cite{MakeupRecommend}.

Although certain image generation models provide the option to generate makeup images, they typically treat makeup as a unified face feature, without offering control over its type and style~\cite{LatentTransformerEdit, HFGANedit, HFGI, diffae}. Recent studies have combined CLIP~\cite{CLIP} or diffusion model~\cite{ho2020denoising} to generate high-quality images with a certain level of makeup control~\cite{CLIP2Protect, clipface, LEDITS, Nulltext, LEDITS++}. However, these language-based makeup image generation methods cannot precisely control makeup details, and often, the same prompt does not produce consistent makeup results.

Our method aims to encode facial makeup to obtain disentangled makeup features. Our makeup encoding can be applied to various applications and expand makeup-related research, enabling new tasks such as enhanced facial image generation with makeup injection and makeup similarity measure.

\subsection{StyleGAN-based Stylized Portrait}
Stylized portrait generation has seen significant advancements~\cite{Toonify, dualstylegan, VToonify, FixtheNoise, DeformToon3d}, particularly through the use of the StyleGAN model~\cite{FFHQ, StyleGAN2} for high-resolution image generation and flexible style control. Approaches like Toonify~\cite{Toonify} fine-tune a pre-trained StyleGAN on cartoon datasets, combining layers from the fine-tuned and original models to generate cartoon-like faces. The pSp method~\cite{pSp} trains an encoder to project real face images into cartoon faces, while DualStyleGAN~\cite{dualstylegan} adds an extrinsic style path for exemplar-based style transfer. StyleGAN-NADA~\cite{StyleGAN-NADA} uses CLIP to guide StyleGAN into new artistic domains without real cartoon datasets, enabling text-driven toonification. StyleGAN inversion techniques~\cite{GANInversion, Image2StyleGAN, pSp, E4e, PTI, HyperInverter, HyperStyle, StyleGANEX, PIE, ReStyle, HFGI} further enhance these capabilities by projecting real face images into StyleGAN's latent space for editing.

In contrast to the challenges faced by stylized portrait methods, such as misalignment caused by artistic styles, our focus is on realistic face images, specifically within two domains: bare-face and makeup-face. We build upon the StyleGAN-based stylized portrait framework~\cite{dualstylegan} and leverage StyleGAN inversion techniques to capture high-dimensional representations of makeup.

\section{Methodology}
\begin{figure}[t]
    \centering
    \includegraphics[width=0.99\linewidth]{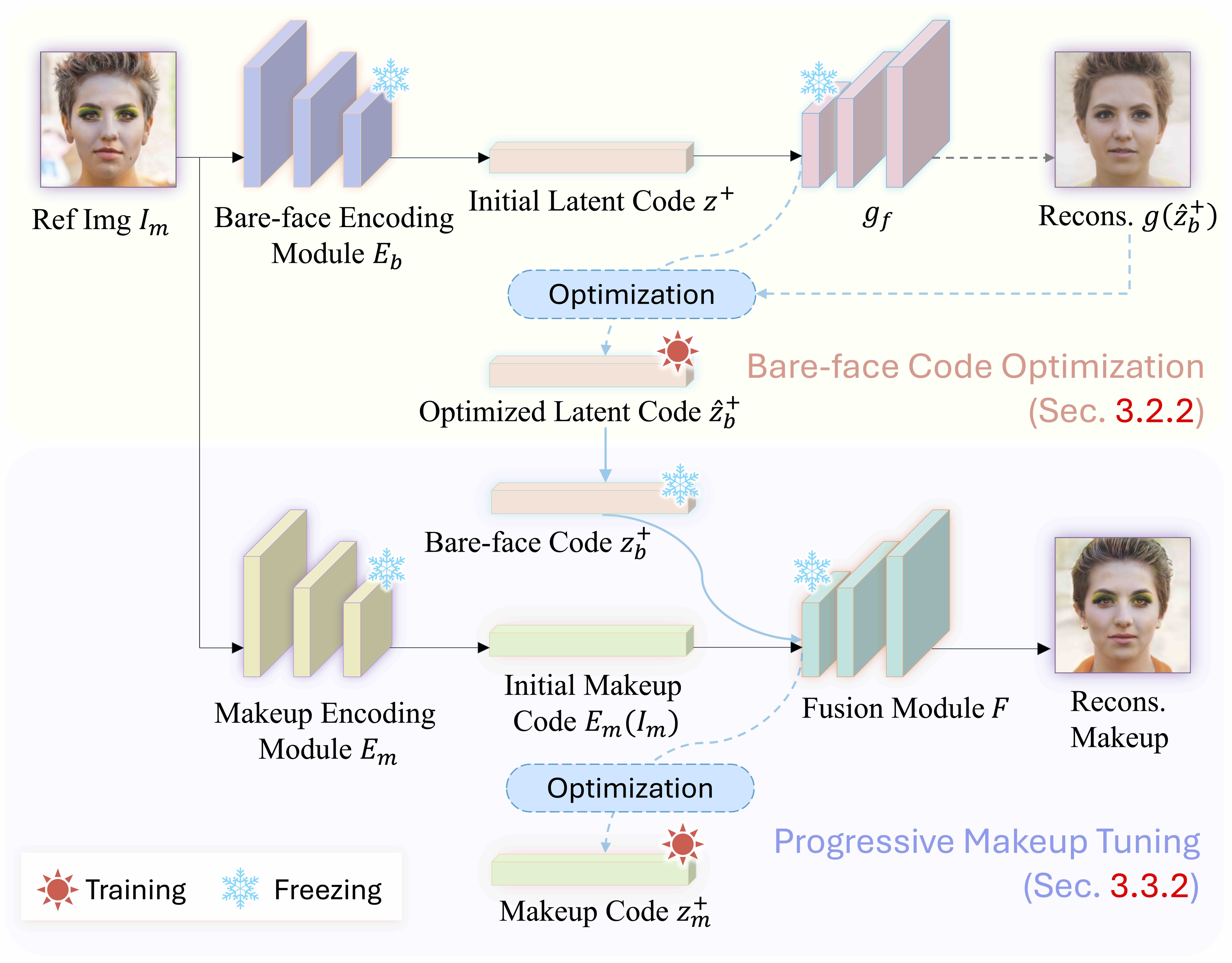}
    \caption{\textbf{The workflow of latent code optimization.} We enhance the encoding of identity information to optimize the bare-face code (see Section~\ref{sec:face code optimization} for details). Subsequently, based on the encoded bare-face code, we use the specially designed objective function to enhance the encoding of makeup details and avoid encoding features unrelated to the makeup, achieving the final makeup encoding (see Section~\ref{sec:PMT} for details).}
    \label{fig:Workflow}
\end{figure}

Our objective is to develop an enhanced model, named BeautyBank (in Section~\ref{sec:BeautyBank}), which is inspired by DualStyleGAN~\cite{dualstylegan}. It encodes makeup to cater to a broader range of makeup-related applications. Our core idea involves incorporating prior knowledge of identity encoding and makeup as supervision, extracting the bare-face code of makeup portraits (in Section~\ref{sec:Bare-face Encoding}). Building on the bare-face code, we employ a progressive fine-tuning strategy specifically designed to optimize makeup codes, preserving more detailed makeup features and reducing unrelated information. (in Section~\ref{sec:Makeup Encoding}). The workflow is illustrated in Fig.~\ref{fig:Workflow}.

\subsection{BeautyBank}
\label{sec:BeautyBank}

Drawing from the network architecture of DualStyleGAN~\cite{dualstylegan}, BeautyBank is designed to extract bare-face and makeup features. \revised{It includes two independent style paths—a bare-face style path and a makeup style path—along with a fusion module $F$.}

\revised{The bare-face style path features a bare-face encoding module $E_b$, constructed based on the pSp encoder~\cite{pSp}, which maps the input facial features to $Z_+$ space. This initial latent code $z^+$ ($z^+=E_b(I)$) is refined to obtain the bare-face code $z_b^+$ ($z_b^+ \in \mathbb{R}^{18 \times 512}$), capturing facial identity and structural features. The input image $I$ can be replaced with the reference makeup image $I_{m}$ if there is no corresponding bare-face image available. Similar to the bare-face style path, the makeup style path incorporates a makeup encoding module, $E_m$, also constructed based on the pSp encoder, which maps makeup features of $I_{m}$ to $Z_+$ space. This results in an initial makeup code, $E_m(I_{m})$, that prepares for subsequent makeup encoding of $I_{m}$. $E_b$ and $E_m$ are both pretrained on the FFHQ dataset. The fusion module $F$ incorporates two mapping networks for $z_b^+$ (the bare-face style path) and $E_m(I_{m})$ (the makeup style path) separately, and a synthesis network to fuse the two latent codes after mapping. This module generates facial images that merge identity features from $z_b^+$ with makeup features from $E_m(I_{m})$. After refining $z_b^+$, we further optimizes the initial makeup code to obtain the final makeup code $z_m^+$ ($z_m^+ \in \mathbb{R}^{18 \times 512}$), which allows for more flexible control over specific makeup features (color and structural features) of the generated image. The style adjustment parameter $w$ ($w \in \mathbb{R}^{18}$), used in $F$, serves as a weight vector for the flexible blending of style features from $z_b^+$ and $z_m^+$, and is preset to 1. When $w$ is set to 0, $F$ degrades to a standard StyleGAN generator $g$ for face generation.}

\begin{figure}[t]
    \centering
    \includegraphics[width=0.99\linewidth]{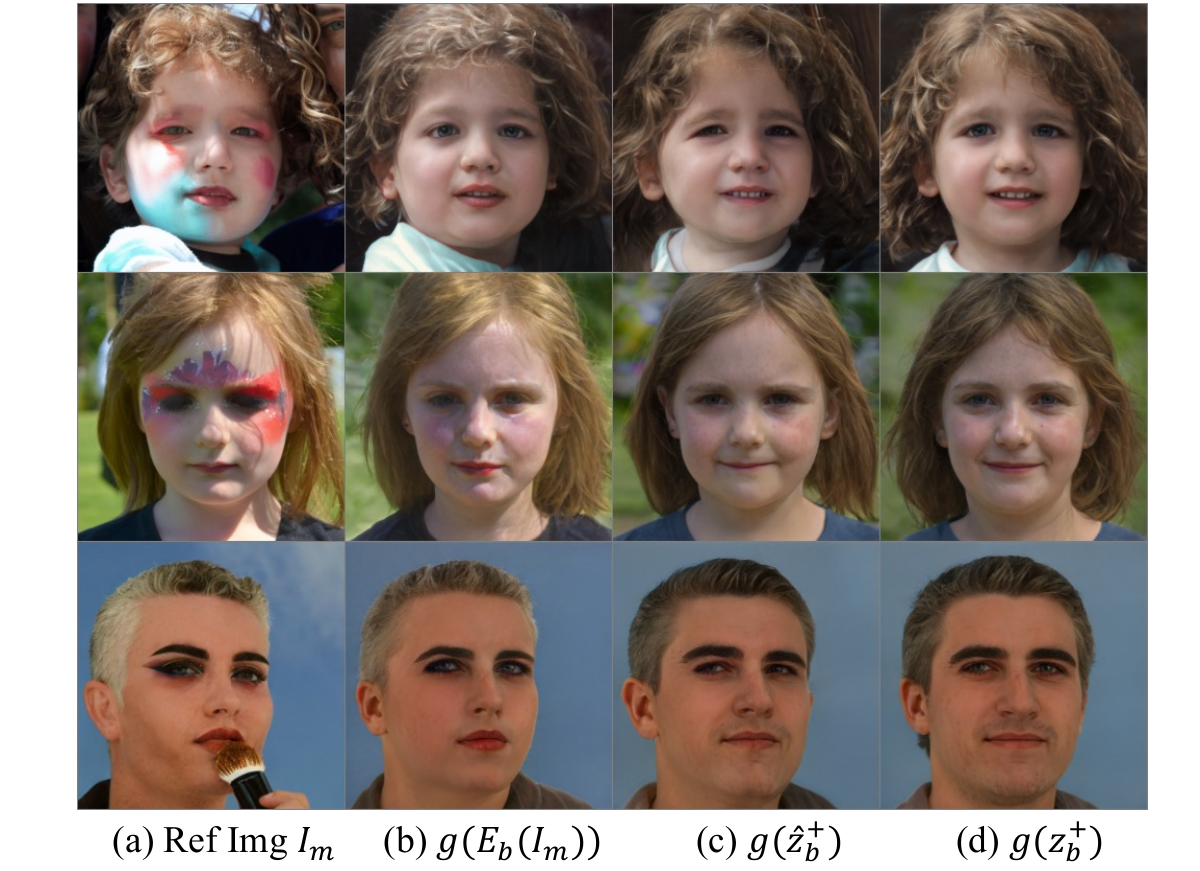}
    \caption{\textbf{Example of bare-face encoding.} Bare-face encoding results from (a) in the preliminary stage (in Section~\ref{sec:Preliminary_destylization}) are shown in (b), while results from bare-face code optimization (in Section~\ref{sec:face code optimization}) are shown in (c) and (d). Bare-face encoding progressively disentangles the makeup information contained in (a) while maintaining consistent identity features.}
    \label{fig:Destylization}
\end{figure}

\subsection{Identity-Optimized Bare-face Encoding}
\label{sec:Bare-face Encoding}

Bare-face encoding aims to disentangle bare face features from \revised{ the reference makeup image $I_{m}$} to guide the subsequent encoding and reconstruction of makeup features. In this section, we first provide a concise introduction to DualStyleGAN~\cite{dualstylegan} in Section~\ref{sec:Preliminary_destylization}, which outlines the methodology for facial destylization. We then present a detailed explanation of our bare-face code optimization method in Section~\ref{sec:face code optimization}.

\subsubsection{Overview of DualStyleGAN}
\label{sec:Preliminary_destylization}

Our bare-face encoding method is an extension of the facial destylization approach proposed in DualStyleGAN~\cite{dualstylegan}. To balance between face realism and fidelity to the portraits, DualStyleGAN proposes a multi-stage destylization method to obtain an intrinsic style code containing facial structure features.

Initially, DualStyleGAN performs the latent initialization of an artistic portrait. Due to the robustness of $Z_+$ space compared to $W_+$ space in handling background details unrelated to the face and distorted shapes, the encoder $E_b$ is utilized to encode the artistic portrait into the $Z_+$ space. 
\revised{Then the initial reconstructed facial image is generated using $g$, which is pretrained on FFHQ.}

Subsequently, the latent codes are refined to better match the facial structures. Although the output of $g$ at this stage, as shown in Fig.~\ref{fig:Destylization} (b), closely resembles the original face due to $g$'s limitations in fully reconstructing the artistic portrait, certain artistic style features are encoded into $Z_+$. \revised{Therefore, DualStyleGAN performs the latent code optimization, and then applies the latent code of $g_f$ back to $g$ to achieve the transfer from the artistic portrait domain to the original face domain. $g_f$ is obtained by further fine-tuning $g$ using makeup images from the BMS dataset.} For more details, please refer to the paper~\cite{dualstylegan}.

\subsubsection{Bare-face Code Optimization}
\label{sec:face code optimization}

\begin{figure}[t]
    \centering
    \includegraphics[width=1.0\linewidth]{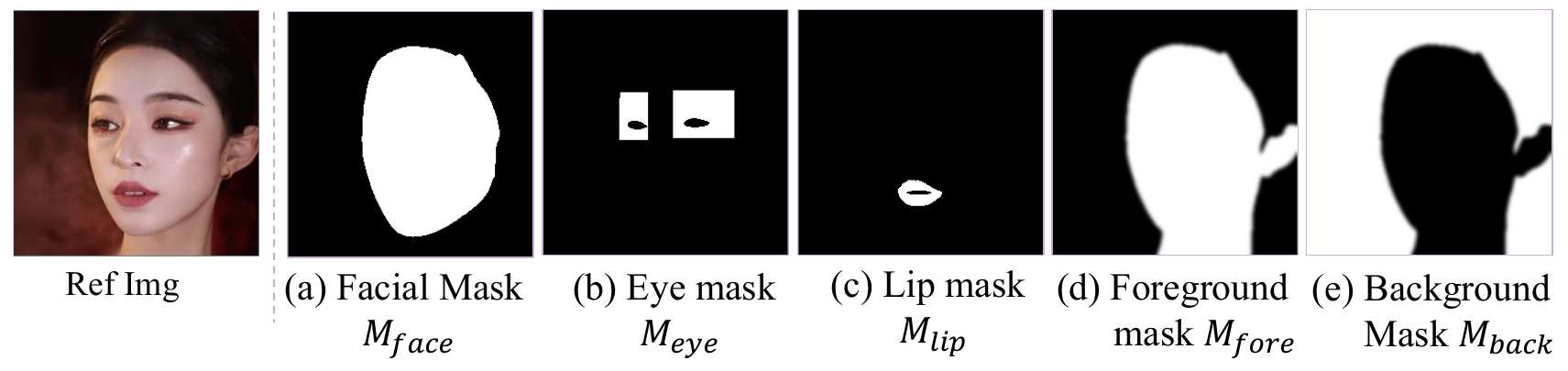}
    \caption{\textbf{Example of the Masks Utilized in BeautyBank.} During Bare-face Code Optimization, the objective function employs mask (a) (in Section~\ref{sec:face code optimization}). Stage 1 of Progressive Makeup Tuning utilizes masks (a), (b), and (c), while Stage 2 employs masks ranging from (a) to (f) (in Section~\ref{sec:PMT}).}
    \label{fig:mask}
\end{figure}

In the process of latent code optimization (as shown in Fig.~\ref{fig:Workflow}), although DualStyleGAN incorporates an identity loss, inconsistencies remain between the identity features of the reconstructed images and the reference makeup (as discussed in Section~\ref{sec:ablation study}). This discrepancy is primarily attributed to the facial recognition model used (ArcFace~\cite{ArcFace}), which does not focus exclusively on the facial region, thereby impacting the accuracy of identity matching. To mitigate the effects of inaccurately encoded bare-face codes on subsequent makeup encoding, we improve the focus on identity features within the facial region during the optimization of the bare-face code. This is achieved by employing a facial mask ($M_{face}$) as shown in Fig.~\ref{fig:mask} (a) and integrating it into the objective function. Specifically, we introduce a facial enhancement loss \revised{$L_{fm}(g_f (z^+), I_{m}, M_{face}) = \| (I_{m} - g_f (z^+)) \odot M_{face} \|_1$}, where \(\odot\) denotes the Hadamard product. This calculates the loss for the facial mask $M_{face}$ region. The full objective function for optimizing the latent encoding is
\revised{
\begin{align*}
L_{b} =  & \lambda_{p_1} L_{perc}(g_f(z^+), I_{m})
+ \lambda_{id} L_{id}(g_f(z^+), I_{m}) \\
& + \lambda_{fm_1} L_{fm}(g_f(z^+), I_{m}, M_{face}) + \| \sigma(z^+) \|_1,
\end{align*}}
where $L_{perc}$ denotes perceptual loss~\cite{PerceptualLoss}, $L_{id}$ is the identity loss~\cite{ArcFace}, and $\sigma(z^+)$ represents the standard error of 18 different 512-dimension vectors in $z^+$, to avoid overfitting during training.  The parameters $\lambda_{p_1}$, $\lambda_{id}$, $\lambda_{fm_1}$ are set to 1, 0.1, and 0.0001, respectively. \revised{By minimizing $L_{b}$, we obtain the optimized latent $\hat{z}_b^+$.
}

Since $g_f$ is a model fine-tuned on the BMS dataset and $g$ is pre-trained on FFHQ, they can be regarded as image generators for the makeup domain and bare-face domain, respectively. Therefore, using the optimized $\hat{z}_b^+$, we obtain $g(\hat{z}_b^+)$ as a bare face image that has removed makeup and retains facial features from \revised{$I_{m}$}. The reconstructed facial image by $g$ is shown in Fig.~\ref{fig:Destylization} (c). Finally, we use the encoder $E_b$ to encode this bare face image, obtaining the bare-face code, $z_b^+ = E_b(g(\hat{z}_b^+))$. Fig.~\ref{fig:Destylization} (d) shows the reconstructed facial image of $z_b^+$.

Furthermore, as the BMS dataset contains paired data of bare faces \revised{$I_{b}$} and makeup \revised{$I_{m}$}, encoding makeup within the BMS dataset simply requires the use of \revised{$z_b^+ = E_b(I_{b})$} to obtain the bare-face code. However, for in-the-wild makeup images that lack paired data, the aforementioned bare-face encoding process is still necessary.

\subsection{Conditional Fine-Tuning Makeup Encoding}
\label{sec:Makeup Encoding}

To obtain a high-dimensional makeup code enriched with detailed makeup information, we perform the pre-training and fine-tuning of BeautyBank, as discussed in Section~\ref{sec:Pre-training and fine-tuning}, and implement the Progressive Makeup Tuning (PMT) strategy for makeup encoding optimization, as outlined in Section~\ref{sec:PMT}.

\subsubsection{Pre-training and fine-tuning of BeautyBank.}
\label{sec:Pre-training and fine-tuning}

Following DualStyleGAN~\cite{dualstylegan}, we conduct pre-training and fine-tuning of the fusion module in BeautyBank to prepare for makeup encoding. To ensure stable and smooth model training, we initially performed the pre-training of the fusion module using the FFHQ dataset. This stage is implemented through color transfer and structural transfer training. Color transfer can stabilize the network parameters without deviating from the original generative space, achieving color migration within the original generative space. Structural transfer involves style mixing operations in the intermediate layers, ensuring the effective capturing and mimicking of detailed structural features while maintaining the color style. \revised{To enable the fusion module to utilize the bare-face code and the makeup code to generate facial images in the makeup domain, we fine-tune the fusion module using facial images from the BMS dataset.} Specifically, we input paired bare-face code $z_b^+$ and \revised{initial makeup code $E_m(I_{m})$} into the fusion module to reconstruct \revised{facial makeup}. The objective function for this stage is
\[
L_{m_1} =  \lambda_{adv} L_{adv} + \lambda_{p_2} L_{perc} + \lambda_{sty} L_{sty} + \lambda_{con_1} L_{con}, 
\]
where parameters $\lambda_{adv}$, $\lambda_{p_2}$, $\lambda_{sty}$, and $\lambda_{con_1}$ are set to 1. $L_{adv}$, $L_{sty}$, and $L_{con}$ denote adversarial loss, style loss, and contextual loss~\cite{ContextualLoss}, respectively. The parameters $\lambda_{adv}$, $\lambda_{p_2}$, $\lambda_{sty}$, and $\lambda_{con_1}$ are set to 1.

\subsubsection{Progressive Makeup Tuning}
\label{sec:PMT}

To better encode essential makeup details and disentangle urelated features, we introduce the Progressive Makeup Tuning (PMT) strategy to optimize the initial makeup code. PMT consists of two stages.

\noindent\textbf{(Stage 1) \revised{Detail-Oriented Latent Optimization:}} \revised{To optimize the makeup detail encoding, we fix the parameters of BeautyBank and fine-tune the makeup code. During this fine-tuning stage, the fusion module in BeautyBank receives paired inputs of the bare-face code and the optimized makeup code. It then reconstructs makeup images to calculate the loss necessary for latent optimization. In the objective function, we incorporate prior knowledge of face parsing to enhance feature extraction in makeup-concentrated regions (overall face, eyes, lips) of facial images.} We apply the objective function
\begin{align*}
L_{m_{2-1}} = & \lambda_{p_3} L_{perc} + \lambda_{con_2} L_{con} + \lambda_{fm_2} L_{fm} \\
& + \lambda_{pm_1} L_{pm} + \lambda_{em_1} L_{em} + \lambda_{lm_1} L_{lm},
\end{align*}
where $L_{pm}$, $L_{em}$, and $L_{lm}$ are the perceptual loss of utilizing the facial mask $M_{face}$ in Fig.~\ref{fig:mask} (a), eye mask $M_{eye}$ in Fig.~\ref{fig:mask} (b), and lip mask $M_{lip}$ in Fig.~\ref{fig:mask} (c). The parameters $\lambda_{p_3}$, $\lambda_{con_2}$, $\lambda_{fm_2}$, $\lambda_{pm_1}$, $\lambda_{em_1}$, and $\lambda_{lm_1}$ are set to 1, 1, 0.0001, 100, 100, 100, respectively.

\noindent\textbf{(Stage 2) \revised{Non-Makeup Features Disentanglement:}} To disentangle makeup-unrelated features (e.g., background, hair color), we further optimize the makeup code. We conduct training using different sources of bare-face code $ z_b^+ $ and makeup code $ z_m^+ $, and replace $ L_{perc} $ with $ \lambda_{pf} L_{pf} + \lambda_{pb} L_{pb} $ in the objective function of the previous stage:
\begin{align*}
L_{m_{2-2}} = & \ \lambda_{pf} L_{pf} + \lambda_{pb} L_{pb} + \lambda_{fm_3} L_{fm} \nonumber + \lambda_{con_3} L_{con} \\
&  + \lambda_{pm_2} L_{pm} \nonumber + \lambda_{em_2} L_{em} + \lambda_{lm_2} L_{lm},
\end{align*}
where $L_{pf}$, $L_{pb}$ represent the perceptual loss of utilizing masks for facial areas, \(M_{fore}\), in Fig.~\ref{fig:mask} (d), and masks for non-facial areas, \(M_{back}\), in Fig.~\ref{fig:mask} (e). In this stage, the output of BeautyBank is a facial image with face and background features from the bare-face code and makeup features from the makeup code. This avoids the inclusion of makeup-unrelated features in the makeup code. The parameters $\lambda_{pf}$, $\lambda_{pb}$, $\lambda_{fm_3}$, $\lambda_{con_3}$, $\lambda_{pm_2}$, $\lambda_{em_2}$, and $\lambda_{lm_2}$ are set to 100, 100, 0.0001, 1, 100, 100, 100, respectively.

Through PMT, BeautyBank achieves bare-face and makeup encoding for 1412 makeup styles. This makeup encoding can be widely applied to various makeup tasks, such as generating faces with specific makeup, makeup transfer, and makeup similarity measure, discussed in Section~\ref{sec:Applications}.

\section{Experiments}

\subsection{Bare-Makeup Synthesis Dataset}
\label{sec:Dataset}

We utilized a pretrained diffusion method LEDITS++~\cite{LEDITS++} to create a large-scale bare-makeup synthesis dataset, Bare-Makeup Synthesis Dataset (BMS). The construction process primarily involves two steps:

First, inspired by Stable-Makeup~\cite{stablemakeup}, we employed GPT-4 to generate 400 style prompts using the template \revised{``make it \{\} makeup"}. However, upon testing these prompts, we found that the generated makeup samples lacked diversity in patterns and colors. Therefore, we used the template \revised{``\{\} makeup with \{\} on the face"} to generate 410 style prompts with GPT-4. To further enhance the diversity of prompts, we constructed 20 color prompts (e.g., Red, Blue, etc.). Ultimately, we created 16,200 prompt pairs by combining the 810 style prompts with the 20 color prompts, which were used to guide the LEDITS++ model in synthesizing makeup data.

Second, we used the FFHQ dataset as the bare skin data to synthesize the corresponding makeup data. For each prompt, we randomly selected 20 facial images from the FFHQ dataset as source images for makeup rendering. 

\begin{figure}[t]
    \centering
    \includegraphics[width=\linewidth]{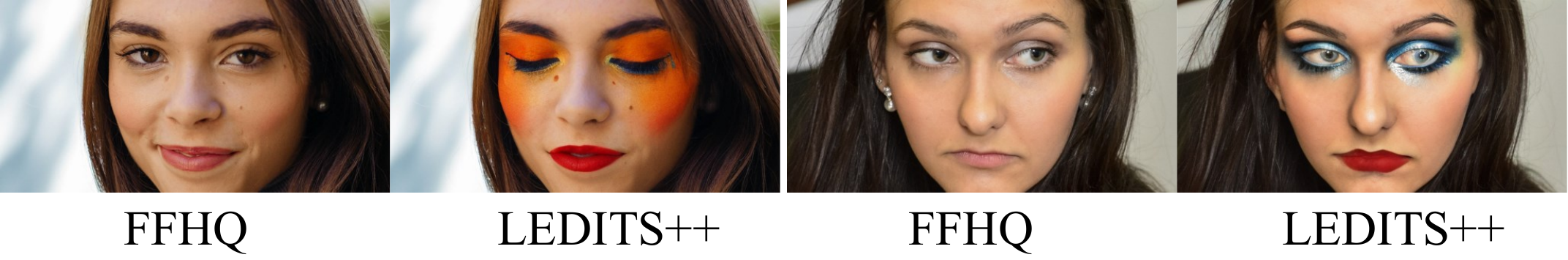}
    \caption{\textbf{Examples of generated makeup images using LEDITS++ with text prompt.} Despite using the same style prompt 'make it fairy makeup' and the color prompt 'Blue', the generated images exhibit markedly different colors and pattern details.}
    \label{fig:ledits}
\end{figure}

Consequently, we constructed the BMS dataset, comprising 324,000 pairs of 512x512 pixel bare-makeup facial images. It should be noted that even when using identical prompts, \revised{LEDITS++} cannot produce consistent makeup results. As shown in Fig.~\ref{fig:ledits}, using the same style prompt \revised{``make it fairy makeup"} and the color prompt \revised{``Blue"}, the generated makeup looks are significantly different. This demonstrates that the prompt code cannot be used as the makeup embedding.

\subsection{Experimental Setup}

We conducted the training of BeautyBank using the PMT strategy. The training was performed on 4 NVIDIA Tesla T4 GPUs, with a batch size of 2 per GPU. For bare-face encoding, the number of training iterations for \(g_f\) was 600, and the number of iterations for optimizing the encoding was 300. In makeup encoding, the number of iterations for each stage of PMT was 300 and 300, respectively. The bare-face images used for training were sourced from the FFHQ dataset, and the makeup images were sourced from the BMS dataset and BeautyFace dataset~\cite{BeautyREC}.

% \subsection{Experimental Results}
Our developed BeautyBank can encode a wide variety of makeup styles. Currently, we have encoded 1412 makeup codes using BeautyBank, all of which are derived from the BMS dataset and BeautyFace dataset. Utilizing these makeup codes, we can perform various makeup-related tasks (in Section~\ref{sec:Applications}), demonstrating the versatility and flexibility of BeautyBank in practical applications. To further expand the application scope of BeautyBank, we plan to encode additional makeup codes in future work to support more diverse makeup image tasks.

\begin{figure*}[t]
    \centering
    \includegraphics[width=0.92\linewidth]{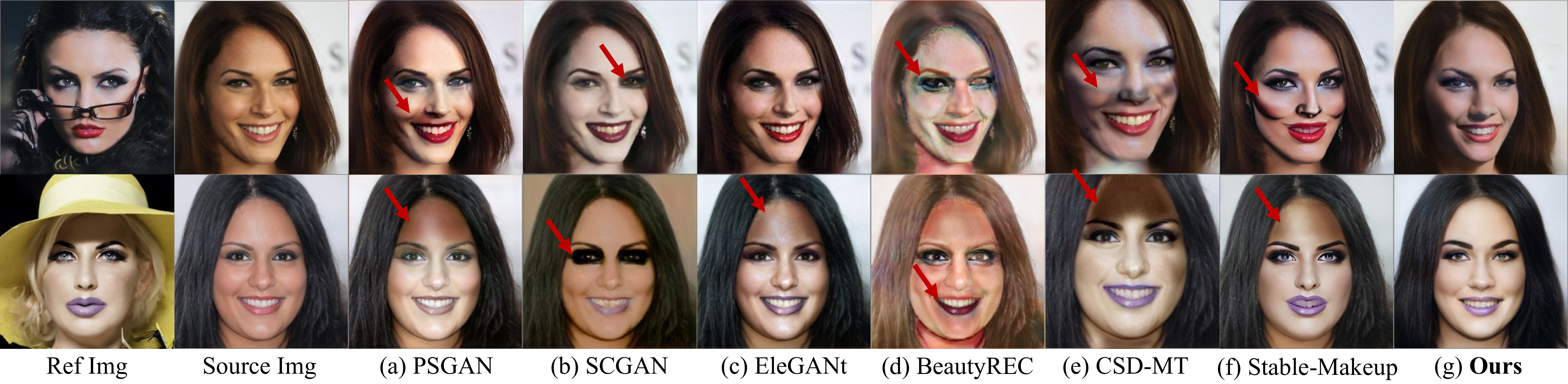}
    \caption{\textbf{Qualitative comparison of different methods.} Our results outperform other methods in terms of color and detail.}
    \label{fig:result_transfer}
\end{figure*}

\begin{figure}[t]
    \centering
    \includegraphics[width=\linewidth]{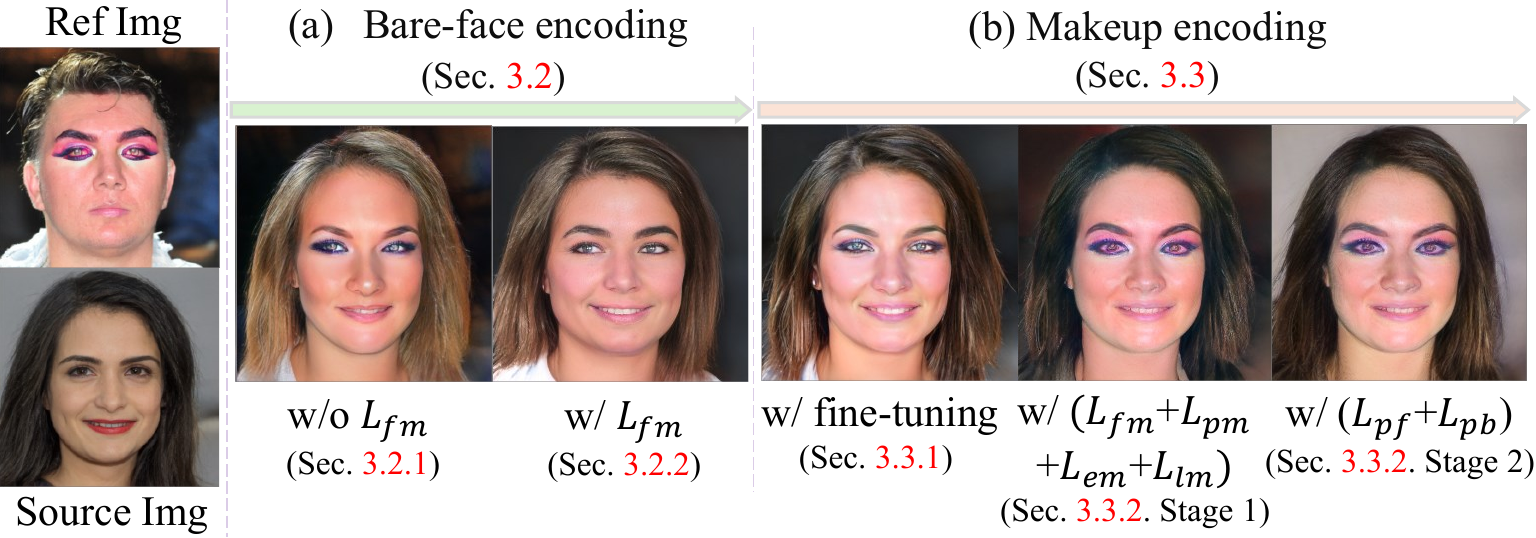}
    \caption{\textbf{Ablation study.} Figure (a) illustrates the ablation study of each stage in bare-face encoding (in Section~\ref{sec:Bare-face Encoding}), while Figure (b) shows the ablation study of each stage in makeup encoding (in Section~\ref{sec:Makeup Encoding}).}
    \label{fig:ablation study}
\end{figure}

\subsection{Comparison with SOTA}

We performed comprehensive comparisons with the most representative makeup transfer algorithms, including PSGAN~\cite{PSGAN} SCGAN~\cite{SCGAN}, EleGANt~\cite{elegant}, BeautyRec~\cite{BeautyREC}, CSD-MT~\cite{CSD-MT}, and Stable-Makeup~\cite{stablemakeup}. As shown in Fig.~\ref{fig:result_transfer}, our results demonstrate more stable performance across various makeup references.

Besides, we conducted a user study to quantitatively evaluate the generation quality and transfer accuracy of different models. We randomly selected 20 pairs of bare-face images from the FFHQ dataset and makeup images from the BMS dataset and BeautyFace dataset, producing 20 makeup transfer result images. A total of 15 participants were asked to evaluate these samples in three aspects: ``visual quality'', ``detail processing'' (the precision of transferred details), and ``overall performance'' (the visual quality, the fidelity of transferred makeup, etc.). Participants were requested to select the best set of results for each aspect. Table~\ref{tab:comparison} shows the results of the user study (ratio (\%) selected as the best). Our BeautyBank outperformed other methods in all aspects. It should be noted that our evaluation data includes reference makeup images with extensive occlusions and shadows, as we aim to evaluate the stability of performance under various conditions.

\begin{table}[t]
\centering
\caption{\textbf{Comparison of different methods based on Quality, Detail, and Overall performance.} Our method received the highest (best) scores across all criteria.}
\label{tab:comparison}
\resizebox{\linewidth}{!}{
\begin{tabular}{@{}lccccccc@{}}
\toprule
Criteria & PSGAN & SCGAN & EleGANt & BeautyRec & CSD-MT & Stable-Makeup & \textbf{BeautyBank} \\
 & \cite{PSGAN} & \cite{SCGAN} & \cite{elegant} & \cite{BeautyREC} & \cite{CSD-MT} & \cite{stablemakeup} & \textbf{(Ours)} \\
\midrule
Quality$\uparrow$   & 0.00\%   & 0.00\%   & 20.00\%   & 0.00\%   & 0.00\%   & 6.67\%   & \textbf{73.33\%} \\
Detail$\uparrow$    & 0.00\%   & 0.00\%   & 40.00\%   & 0.00\%   & 0.00\%   & 13.33\%  & \textbf{46.67\%} \\
Overall$\uparrow$   & 0.00\%   & 0.00\%   & 26.67\%   & 0.00\%   & 0.00\%   & 6.67\%   & \textbf{66.67\%} \\
\bottomrule
\end{tabular}
}
\end{table}

\subsection{Ablation Study}
\label{sec:ablation study}

This section demonstrates the effectiveness of bare-face encoding and makeup encoding by showcasing results on makeup image generation and makeup transfer tasks. As shown in Fig.~\ref{fig:ablation study}, our results demonstrate more stable performance across various makeup references.

\noindent\textbf{Bare-face encoding:} Fig.~\ref{fig:ablation study} (a) shows the performance in the makeup transfer task before and after adding \(L_{fm}\) during the optimization stage. Without \(L_{fm}\), the loss of identity features is more pronounced under the same number of iterations. Additionally, it is worth noting that the makeup transfer results shown in this section are all generated by BeautyBank after completing the stage 1 of PMT.

\noindent\textbf{Makeup encoding:}  Fig.~\ref{fig:ablation study} (b) presents the results from BeautyBank training along with stages 1 and 2 of PMT in the makeup transfer task. With the addition of the detail-enhanced objective function, BeautyBank can fully transfer the color and pattern of the makeup. After further latent optimization, as the makeup code contains fewer makeup-unrelated features, BeautyBank can better preserve the hair and background color of the source image.

\begin{figure}[t]
    \centering
    \includegraphics[width= \linewidth]{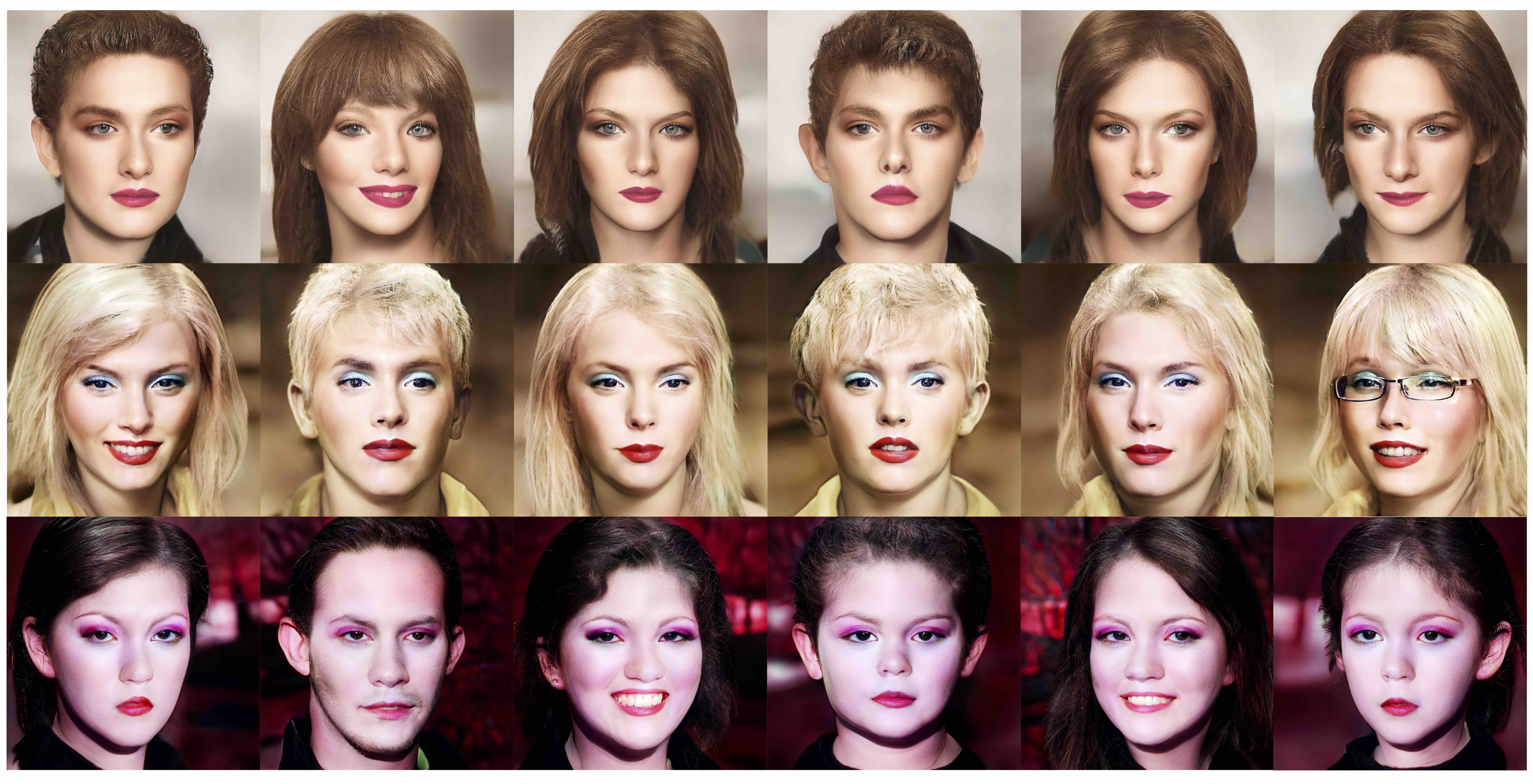}
    \caption{\textbf{Examples of makeup facial generation with makeup injection.} We replace the bare-face code with random Gaussian noise as input to BeautyBank, generating facial images with the same makeup but varying in gender, expressions, hairstyles, and face shapes.}
    \label{fig:facegeneration}
\end{figure}

\begin{figure}[t]
    \centering
    \includegraphics[width=  \linewidth]{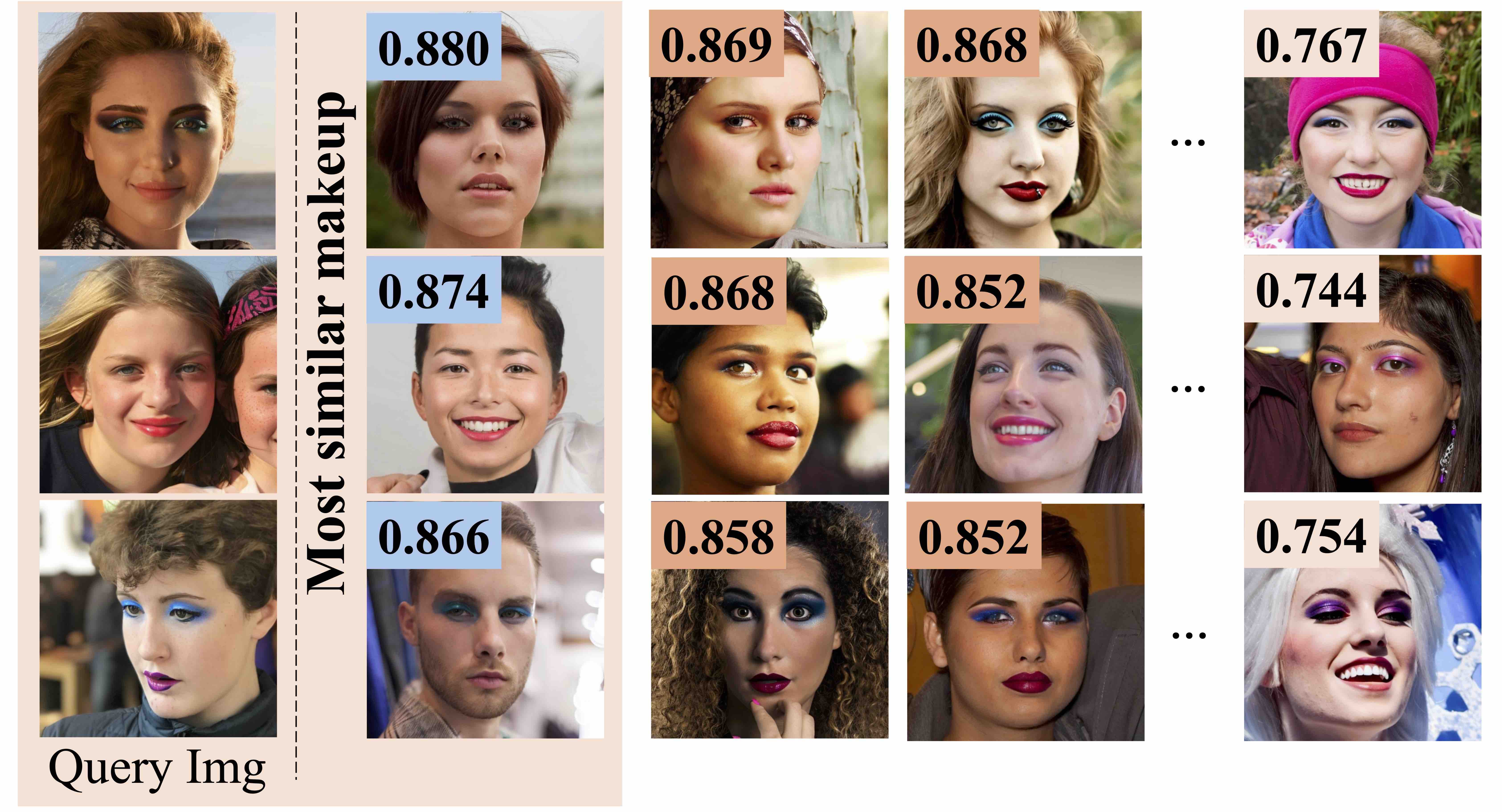}
    \caption{\textbf{Examples of makeup similarity measure with reference makeup.} By searching the encoded makeup database and calculating the cosine similarity with the makeup code of the query image, we can identify the makeup style most similar to the query image.}
    \label{fig:CosineSimilarity}
\end{figure}

\begin{figure}[t]
    \centering
    \includegraphics[width= 0.95\linewidth]{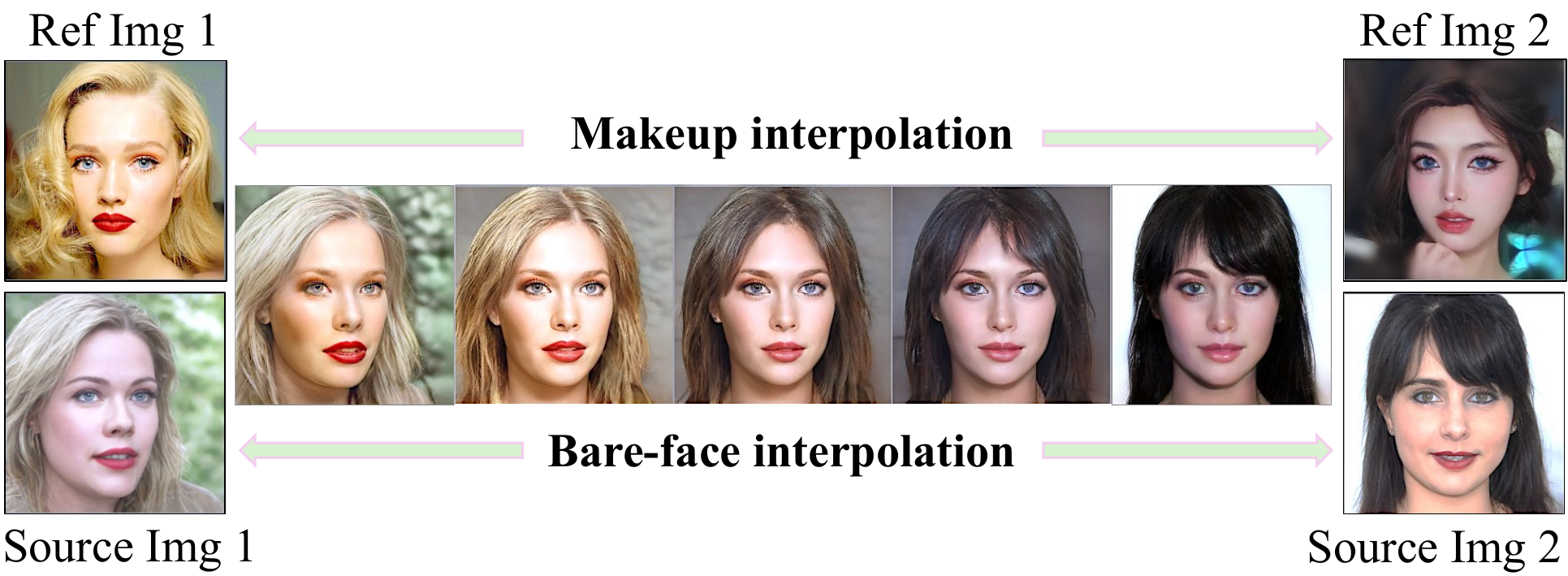}
    \caption{\textbf{Makeup interpolation application.} BeautyBank can separately encode the bare-face code \revised{(Source Img 1 and 2)} and the makeup code \revised{(Ref Img 1 and 2)}, supporting interpolation between different sets of bare-face and makeup images.}
    \label{fig:interpolation}
\end{figure}

\begin{figure}[t]
    \centering
    \includegraphics[width= \linewidth]{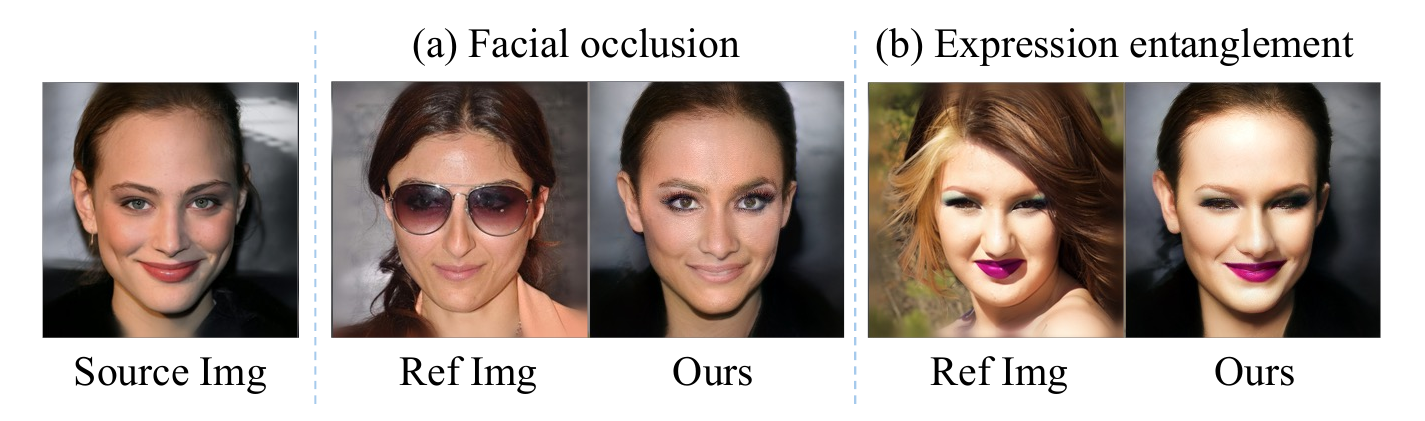}
    \caption{\textbf{Limitations of BeautyBank.} The makeup images generated by Beauty perform poorly in cases of extensive facial occlusion, or exhibit entangled expression information due to the limitations of the image encoder.}
    \label{fig:limitations}
\end{figure}

\section{Applications}
\label{sec:Applications}

To explore the effectiveness of our method, we evaluated our makeup encoding on several makeup-related applications.

\noindent\textbf{Makeup facial generation with makeup injection:} We randomly selected several sets of encoded makeup codes, and for each makeup code, we generated random Gaussian noises to replace the bare-face code. Subsequently, we used the fusion module of BeautyBank for facial image generation. Fig.~\ref{fig:facegeneration} illustrates the results of our facial image generation. The figure indicates that by altering the input random noise, we can generate faces with various expressions, poses, genders, and hairstyles, while retaining the specified makeup. 

\noindent\textbf{Makeup similarity measure:} As shown in Fig.~\ref{fig:CosineSimilarity}, by calculating and ranking the cosine similarity between makeup codes, we can retrieve similar makeup styles from the encoded makeup database. The examples shown are from the 1412 encoded makeup styles. With more makeup encoded, more accurate and similar results can be obtained.

\noindent\textbf{Makeup transfer:} As shown in Fig.~\ref{fig:result_transfer}, BeautyBank can perform makeup transfer by utilizing the bare-face code from the source image and the makeup code from the reference makeup image. The generated images using BeautyBank are overall more natural and realistic, with rich colors and detailed features in the makeup.

\noindent\textbf{Makeup removal:} As shown in Fig.~\ref{fig:Destylization} (d), BeautyBank can generate bare-skin facial images with preserved identity features by performing bare-face encoding of the input makeup image \revised{$I_{m}$}. 

\noindent\textbf{Makeup interpolation:} Demonstrated in Fig.~\ref{fig:teaser} (f) and \ref{fig:interpolation}, since BeautyBank includes two style paths, we can achieve seamless interpolation between different source images and reference makeup styles by interpolating between the bare-face codes or between the makeup codes.

\section{Conclusion}

In this study, we introduced BeautyBank, a novel makeup encoding approach that significantly expands the application possibilities in the field of makeup. We also developed the Bare-Makeup Synthesis Dataset (BMS) and the Progressive Makeup Tuning (PMT) strategy, which enhance the extraction and refinement of makeup codes. Extensive empirical testing confirms that our approach not only improves the adaptability of makeup tasks but also opens up new avenues for innovative applications such as makeup injection and similarity measure. We believe these advancements set a new standard for future research and applications in makeup-related technologies.

As illustrated in Fig.~\ref{fig:limitations}, although our model demonstrates robustness in accurately encoding makeup from reference images with partial facial occlusions, significant occlusions can lead to incorrect encoding in these areas. \revised{Moreover, accurately estimating natural skin tone from images with makeup presents challenges, primarily because most makeup applications include a foundation layer. Consequently, our methodology assumes that the input facial images already have foundation applied. Additionally, due to the variability in iris color—which may be natural or altered by cosmetic lenses—we do not categorize it as unrelated to makeup. Therefore, both the foundation color and iris color in our generated results are closely aligned with the reference makeup. Furthermore, the accuracy of our makeup encoding process, which utilizes the pSp encoder~\cite{pSp}, is constrained by the capabilities of this model. Challenges such as effectively disentangling facial expressions or avoiding identity shifts during the encoding process may occur. Moving forward, we plan to explore the use of higher-quality facial encoders and develop specialized methods aimed at more effectively disentangling expressions while preserving identity features to overcome these limitations.}

%%%%%%%%% REFERENCES
{\small
\bibliographystyle{ieee_fullname}
\bibliography{egbib}
}

\clearpage

%%%%%%%%% TITLE - PLEASE UPDATE
\clearpage
\twocolumn[ 
\begin{center}
    {\LARGE \textbf{Supplementary Material}}
\end{center}
\vspace{1em} 
]

\appendix

%%%%%%%%% BODY TEXT - ENTER YOUR RESPONSE BELOW

\begin{figure*}[t]
    \centering
    \includegraphics[width=0.99\linewidth]{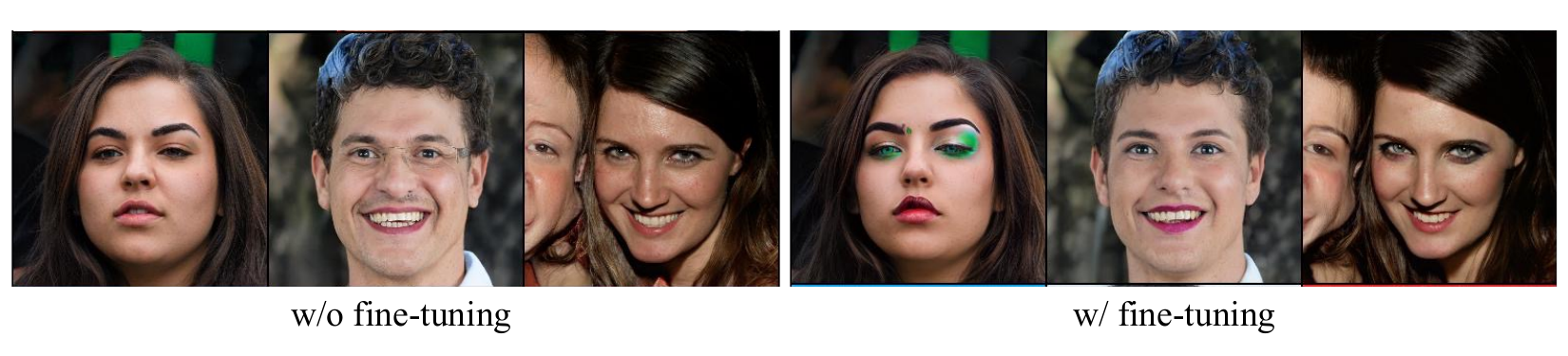}
    \caption{Examples of generated images by the \(g\) and \(g_f\) networks before and after fine-tuning.}
    \label{fig:stylegan}
\end{figure*}

In this supplemental material, we first provide additional training details in Section \ref{supp:train_details}. Then, we present more details about the dataset in Section \ref{supp:dataset} and \ref{supp:code}, as well as additional experiments.

\section{Training Details}
\label{supp:train_details}

\subsection{Bare-Face Encoding}

\revised{Following the training method for $g'$ in DualStyleGAN, after initially training the generator \( g \) with the FFHQ dataset, we performed finetuning on \( g_f \) using images from our BMS dataset. This approach enabled \( g_f \) to effectively generate images within the makeup domain.} The outputs from \( g \) and \( g_f \) are illustrated in Fig.~\ref{fig:stylegan}, which demonstrates the network's enhanced ability to generate various makeup colors and patterns.

Subsequently, in the bare-face code optimization (in Section \textcolor{red}{3.2.2}), we fixed the parameters of \( g_f \) and used the 1,412 makeup images as label images. The initial latent code, \( z^+ \) {(\( z^+ = \revised{E_b (I_{m})} \in \mathbb{R}^{18 \times 512} \)), was optimized using facial enhancement loss of the reconstructed facial images and the label images. Additionally, in the facial enhancement loss, a facial mask was derived by performing face segmentation on the 1,412 makeup images using a face-parsing method \cite{BiSeNet}. This mask exclusively contains the facial region of the images to improve the learning of identity features.

\subsection{Makeup Encoding}

\revised{In Makeup Encoding, the makeup encoding module, \( E_m \), and the bare-face encoding module, \( E_b \), share identical network architectures and parameters. The latent code output by \( E_m \) prepares for subsequent makeup style encoding.}

We utilized FFHQ dataset for pre-training BeautyBank, and fine-tuned BeautyBank with 130 selected images from 1,412 images, as detailed in Section \textcolor{red}{3.3.1}. Subsequently, in the fine-tuning of the makeup code (Section \textcolor{red}{3.3.2}), stage 1 involved computing the objective function using facial makeup images reconstructed from the initial makeup code \revised{\(E_m(I_{m})\)} and using 1,412 images as label images. The objective function includes \(L_{pm}\), \(L_{em}\), and \(L_{lm}\), all applying the Hadamard product for perceptual loss. Eye and lip masks were obtained using a face-parsing method \cite{BiSeNet}. The eye mask includes areas corresponding to the bounding rectangles of both sets of eyes and eyebrows. To encompass the richly detailed makeup region beneath the eyes, we extended the bounding rectangles downward by 1.3 times their height and also excluded areas within the eye socket and any part of the rectangle extending beyond the face. The lip mask solely includes the areas of the upper and lower lips, excluding the interior of the mouth. In this stage, the first 7 rows of \(z_m^+\) had a learning rate of 0.005, while the last 11 rows of \(z_m^+\) had a learning rate of 0.1. In stage 2, we utilized a foreground mask including the face and neck areas and a background mask for other areas. Given the changes in features like facial shape during reconstruction, to ensure a smooth transition between the face and other parts, we applied Gaussian blurring with a kernel size of 11 to both the foreground and background masks. In this stage, we used label images from 1412 makeup images for \(L_{pf}\), and label images from reconstructed images using bare-face codes for \(L_{pb}\). During training of this stage, the first 7 rows of \(z_m^+\) had a learning rate of 0.005 or 0.001, while the last 11 rows of \(z_m^+\) used learning rates of 0.01 or 0.005. Additionally, in the makeup transfer task, the source image can be used instead of the reconstructed image as the label for \(L_{pb}\) to achieve better performance.

\section{\revised{More Information on our BMS Dataset}}
\label{supp:dataset}

\revised{We will publicly release the Bare-Makeup Synthesis Dataset. We analyzed 324,000 makeup images of 512x512 resolution from the BMS dataset, synthesized based on the FFHQ dataset, using an open-source gender-and-age detector~\cite{Gender-and-Age-Detection}. As shown in Fig.~\ref{fig:dataset}, the proportion of male and female images is 63.71\% and 36.29\%, respectively. The images are distributed across the following age ranges: 0-20, 21-32, 33-53, and 54-100, with respective proportions of 44.48\%, 36.91\%, 16.62\%, and 1.99\%. It should be noted that our analysis includes only those images that were successfully detected by the gender-and-age detector, due to occasional failures in face detection. This demonstrates the diversity in gender and age of the facial images within the BMS dataset. Additionally, examples of paired bare-face and makeup images in our BMS dataset can be seen in Fig.~\ref{fig:dataset_pair}.
}

\section{Encoded Makeup Codes}
\label{supp:code}

We carefully selected 1412 makeup data from our BMS dataset and BeautyFace~\cite{BeautyREC} for encoding. As shown in Fig.~\ref{fig:show_data}, these encoded makeup data are rich and diverse in color, texture, and pattern. We aligned all the makeup data based on facial landmarks following the FFHQ~\cite{FFHQ}. In future work, we plan to select more high-quality makeup images for encoding to expand the application of our method across various makeup scenarios.

\begin{figure*}[t]
    \centering
    \includegraphics[width=0.6\linewidth]{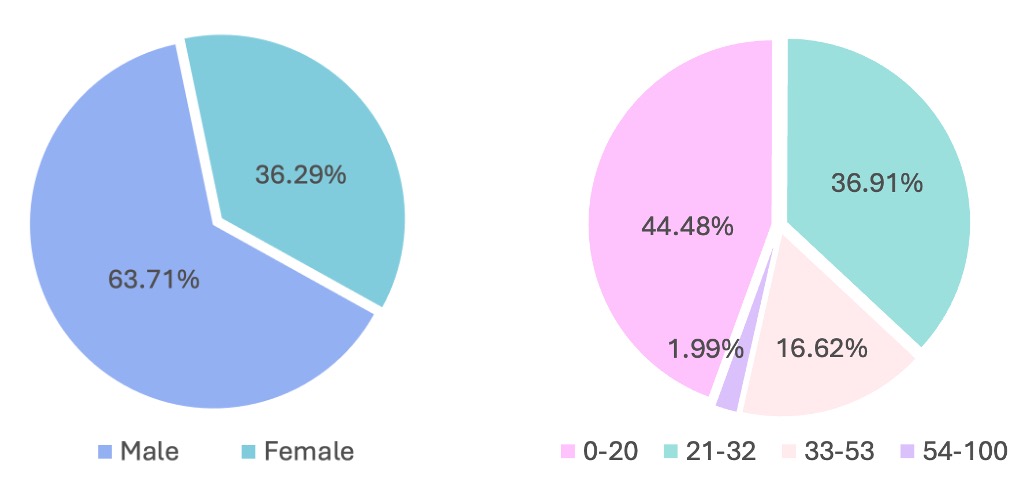}
    \caption{\revised{Gender and age distribution of our BMS dataset.}}
    \label{fig:dataset}
\end{figure*}

\begin{figure*}[t]
    \centering
    \includegraphics[width=0.9\linewidth]{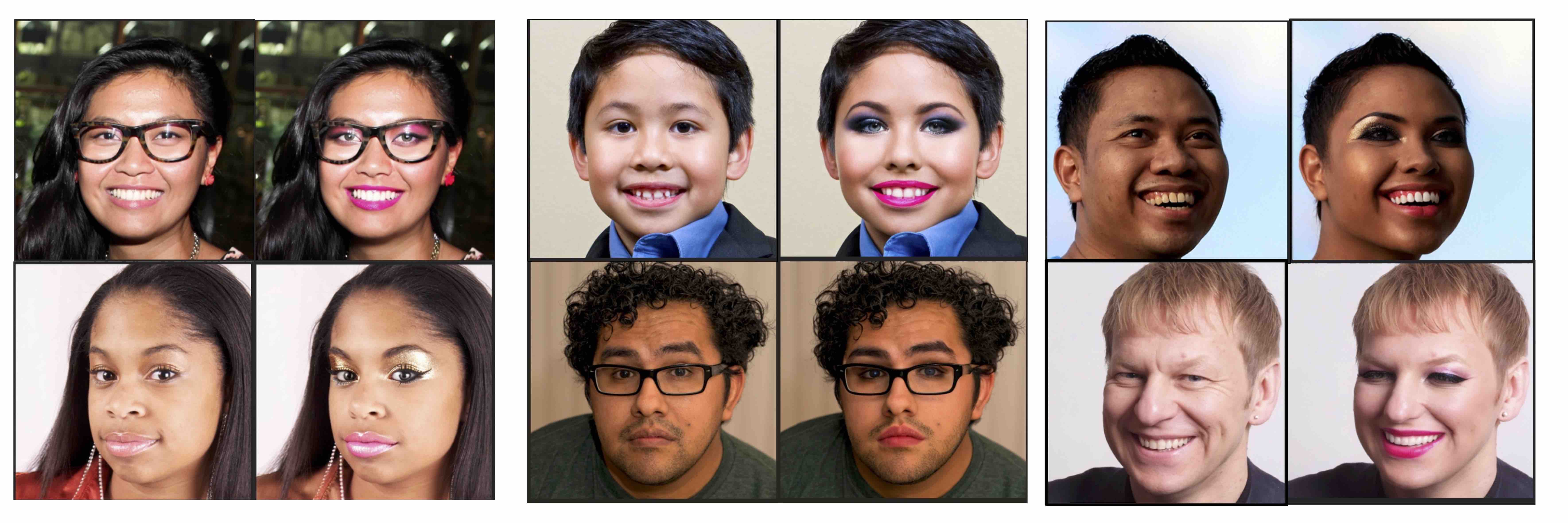}
    \caption{\revised{Paired bare-face and makeup images in our BMS dataset.}}
    \label{fig:dataset_pair}
\end{figure*}

\begin{figure*}[t]
  \centering
  %\fbox{\rule{0pt}{0.5in} \rule{0.9\linewidth}{0pt}}
  \includegraphics[width=0.99\linewidth]{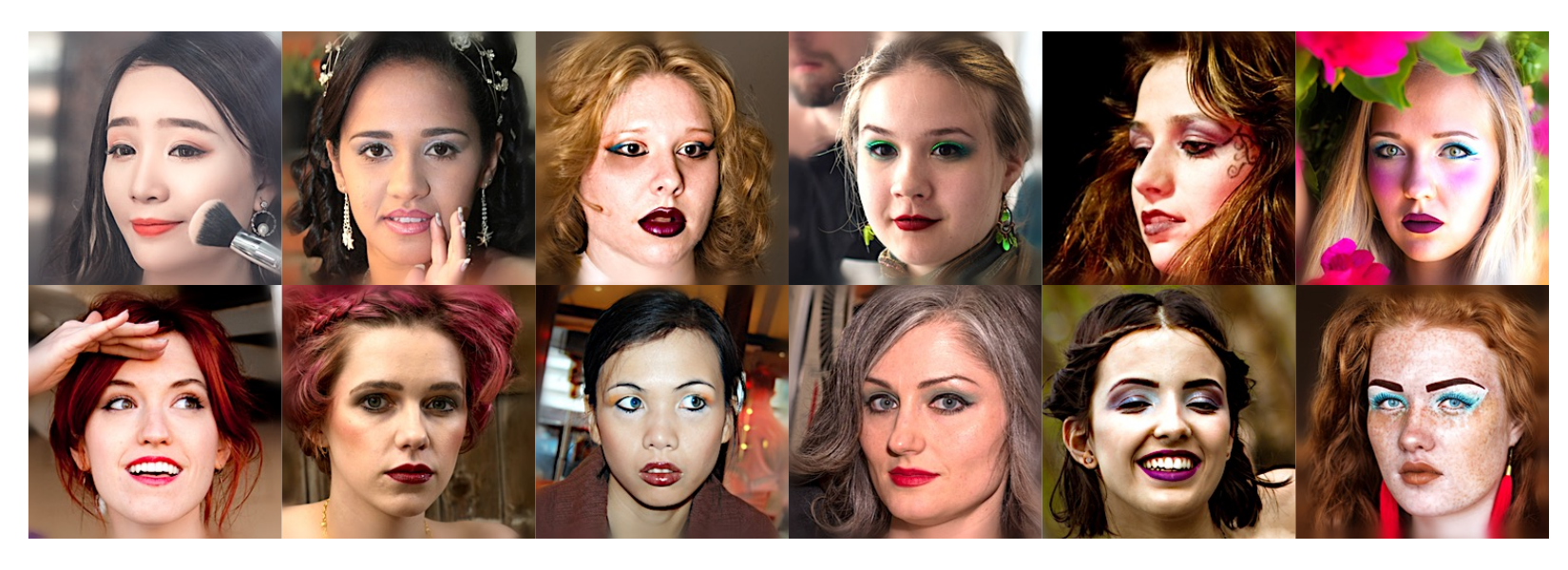}
   \caption{Examples of selected 1412 makeup images.}
   \label{fig:show_data}
\end{figure*}

\section{More Makeup Transfer Results}
\label{supp:transfer}

We provide additional results that highlight the robustness and superiority of our BeautyBank in the task of makeup transfer. As shown in Fig.~\ref{fig:transfer_supp}, BeautyBank successfully generates makeup images that preserve the identity features of the source image while faithfully transferring the makeup attributes from the reference image, including its colors, textures, and detailed patterns. These generated images demonstrate the effectiveness of our approach.

\section{Ablation Study of Weights}
\label{supp:weights}

Our method enables editing of generated images by adjusting 18 weights between the makeup code and bare-face code, each ranging from 0 to 1. We randomly selected three encoded makeups and conducted makeup transfer on bare-faced photos by setting these 18 weights to 0.2, 0.4, 0.6, 0.8, and 1, respectively. As demonstrated in Fig.~\ref{fig:weight}, we can progressively increase the weights to generate makeup results that more closely match the reference makeup in terms of color, texture, and pattern.

\section{\revised{Comparative analysis with different masks}}
\label{supp:mask}

\revised{Our method enables the editing and control of makeup by modifying the masks for facial areas, \(M_{\text{fore}}\), and for non-facial areas, \(M_{\text{back}}\), as mentioned in Section 3.3.2. To prevent the influence of makeup style on the iris during makeup transfer, we utilized the foreground mask (a) and background mask (b) shown in Fig.~\ref{fig:iris} for obtaining the transferred images. As illustrated in Fig.~\ref{fig:iris}, the (d) makeup transfer results (d) using masks (a) and (b) exhibit an iris color that is much closer to the iris color in the source image compared to the image (c) without using the masks. This experiment demonstrates that our method supports flexible control over makeup transfer results through the editing of masks, thereby ensuring more natural and precise makeup application in targeted regions.}

\begin{figure*}[t]
    \centering
    \includegraphics[width=0.99\linewidth]{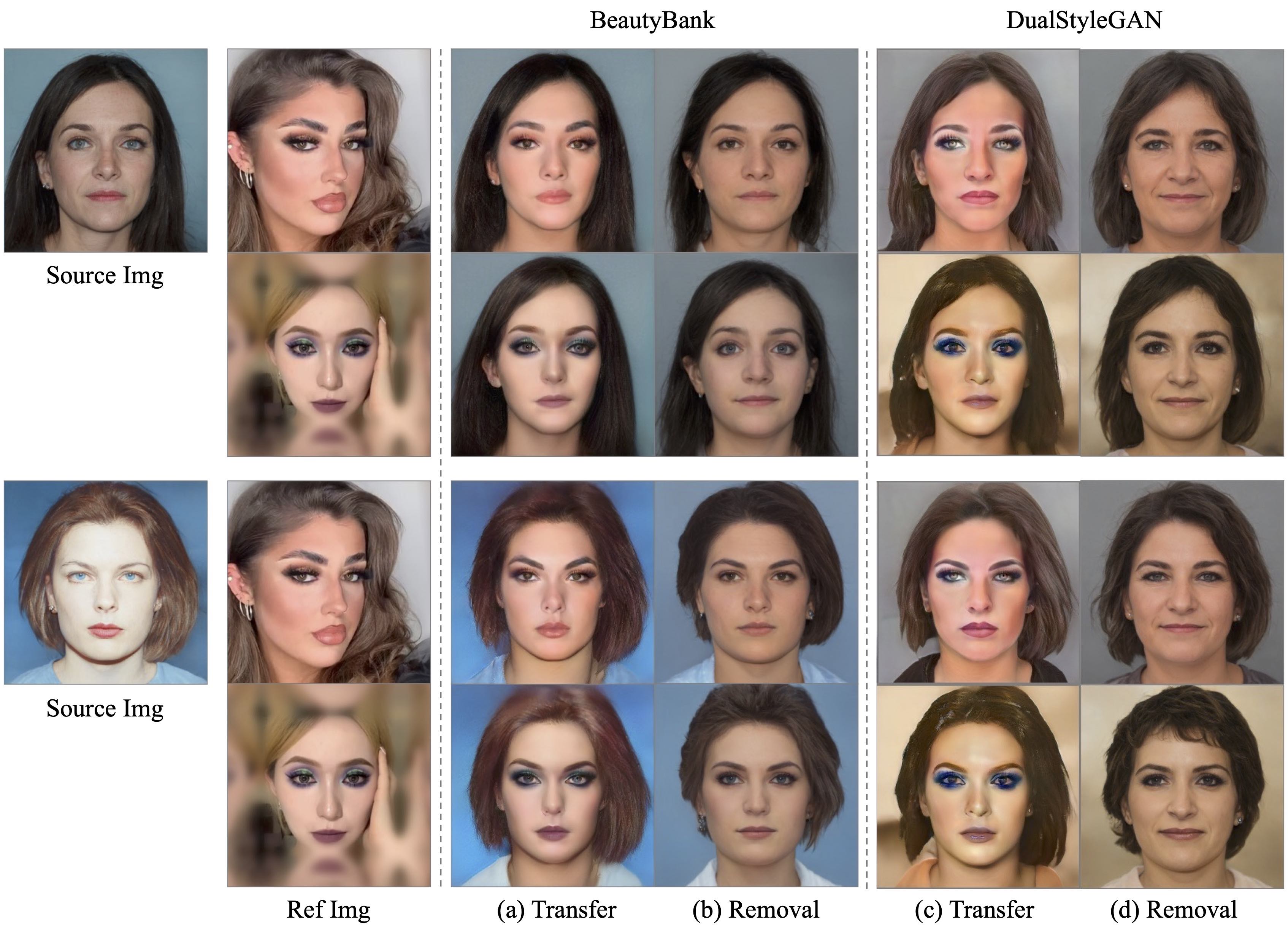}
    \caption{\revised{Comparative analysis with BeautyBank and DualStyleGAN.}}
    \label{fig:compare}
\end{figure*}

\begin{table*}[h]
    \centering
    \caption{\revised{Quantitative comparison of identity for the cycle self-reconstruction experiment. The first group compares the makeup transfer results with the source image, while the second group compares the results after makeup removal (following the transfer) with the source image, using ArcFace~\cite{ArcFace} and BlendFace~\cite{blendface} cosine similarity metrics.}}
    \begin{tabular}{|c|c|c|c|c|}
        \hline
        \multirow{2}{*}{Method} & \multicolumn{2}{c|}{\textit{Transfer}} & \multicolumn{2}{c|}{\textit{Removal}} \\
        \cline{2-5}
        & ArcFace $\uparrow$ & BlendFace $\uparrow$ & ArcFace $\uparrow$ & BlendFace $\uparrow$ \\
        \hline
        DualStyleGAN & 0.174 & 0.148 & 0.084 & 0.130 \\
        \hline
        \textbf{BeautyBank (Ours)} & \bf{0.177} & \bf{0.164} & \bf{0.206} & \bf{0.188} \\
        \hline
    \end{tabular}
    \label{tab:arcface_blendface}
\end{table*}

\section{\revised{Comparison with DualStyleGAN}}
\label{supp:compare_with_dualstylegan}

\revised{We conducted comparative analyses between BeautyBank and DualStyleGAN using a self-reconstruction approach. Fig.~\ref{fig:compare} shows the makeup transfer results ((a) and (c)) using both source and reference images, and makeup removal results ((b) and (d)) after bare-face encoding of (a) and (c), for both BeautyBank and DualStyleGAN. These results demonstrate that our method preserves identity more effectively and retains finer makeup details, while successfully disentangling information irrelevant to the makeup, such as the background.}

\revised{Additionally, our method exhibits superior maintenance of facial identity throughout the makeup transfer and the subsequent removal process. We evaluated the transferred images from both BeautyBank and DualStyleGAN against the source images using ArcFace~\cite{ArcFace} and BlendFace~\cite{blendface} cosine similarity metrics, with the results shown in the '\textit{Transfer}' column of Table~\ref{tab:arcface_blendface}. Similarly, images resulting from the self-reconstruction with BeautyBank and DualStyleGAN were also evaluated against the source images, as shown in the '\textit{Removal}' column. The results indicate that our method more effectively maintains facial identity features during both makeup transfer and self-reconstruction phases. It should be noted that since our method primarily aims to reconstruct bare facial images during the bare-face encoding stage, areas not associated with the face do not require high fidelity reconstruction in our tasks.}

\begin{figure*}[t]
    \centering
    \includegraphics[width=0.99\linewidth]{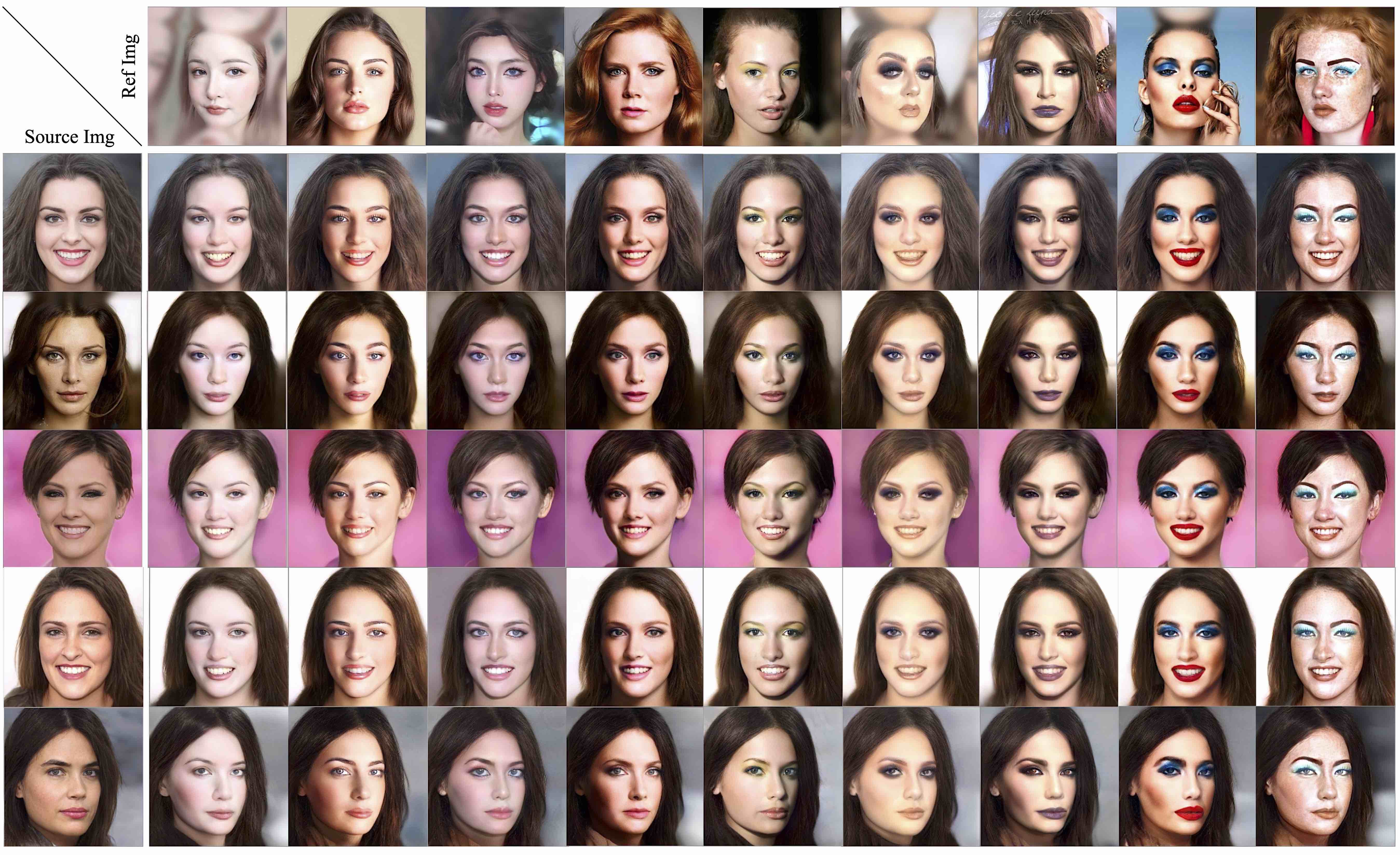}
    \caption{\revised{More results of our BeautyBank in the makeup transfer task.}}
    \label{fig:transfer_supp}
\end{figure*}

\begin{figure*}[t]
    \centering
    \includegraphics[width=0.99\linewidth]{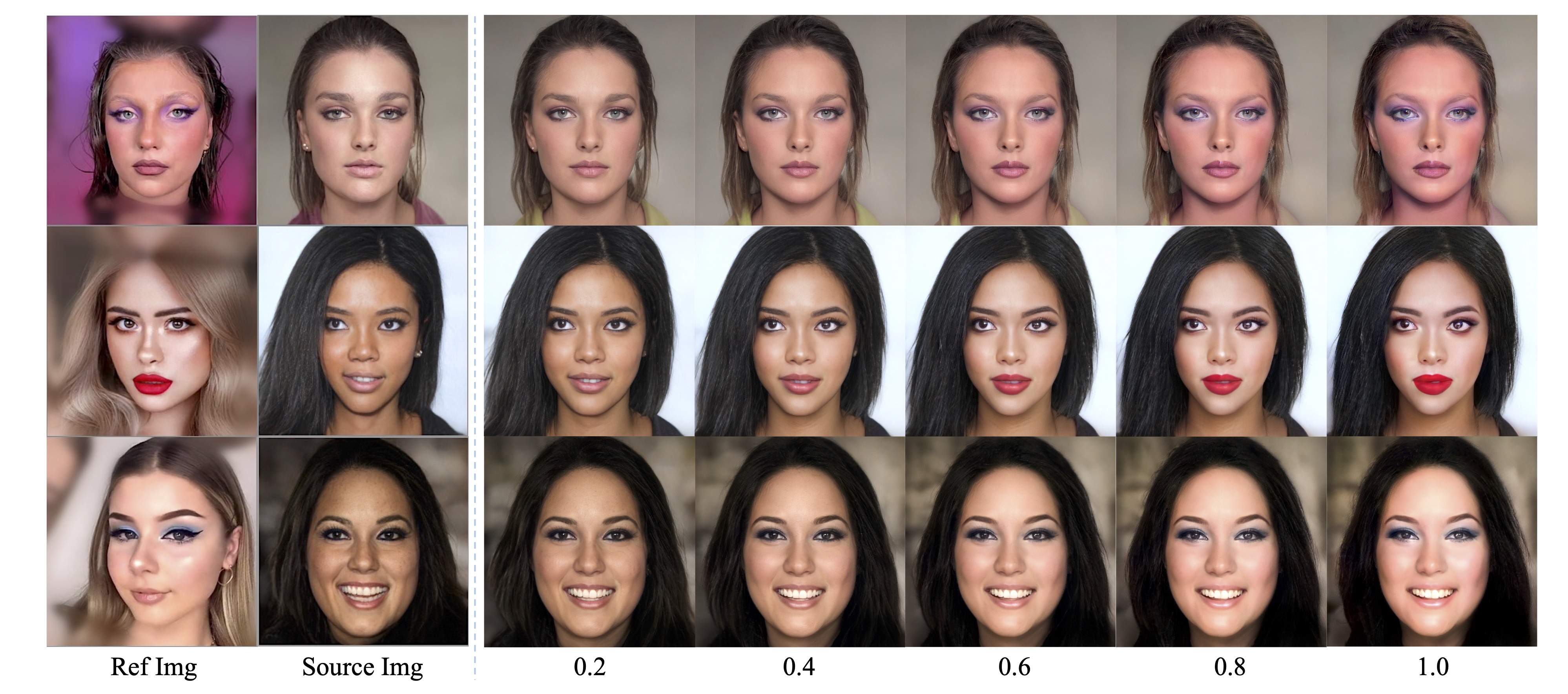}
    \caption{Ablation study of makeup transfer results with different weights.}
    \label{fig:weight}
\end{figure*}

\begin{figure*}[t]
    \centering
    \includegraphics[width=0.99\linewidth]{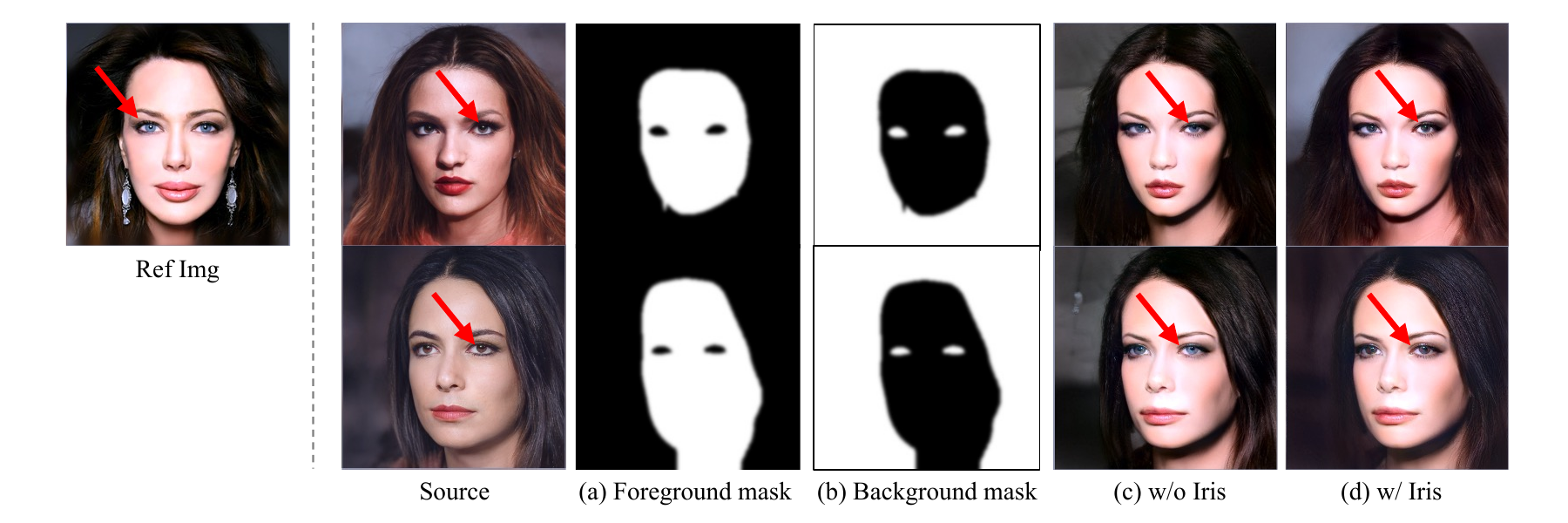}
    \caption{\revised{Comparative analysis of results using different masks to determine and control the color of iris. By utilizing the iris region mask, we can determine whether the iris color comes from the source or the reference image.}}
    \label{fig:iris}
\end{figure*}

\end{document}